\documentclass[10pt,twocolumn,letterpaper]{article}

\usepackage[pagenumbers]{cvpr} 
\usepackage{graphicx}
\usepackage{amsmath}
\usepackage{amssymb}
\usepackage{booktabs}
\usepackage{enumitem}
\usepackage[accsupp]{axessibility}

\usepackage{multirow}

\usepackage[pagebackref,breaklinks,colorlinks]{hyperref}

\usepackage[capitalize]{cleveref}
\crefname{section}{Sec.}{Secs.}
\Crefname{section}{Section}{Sections}
\Crefname{table}{Table}{Tables}
\crefname{table}{Tab.}{Tabs.}

\def\cvprPaperID{5} 
\def\confName{CVPR}
\def\confYear{2023}

\begin{document}

\title{$R^{2}$Former: Unified $R$etrieval and $R$eranking Transformer for Place Recognition}
\author{Sijie Zhu$^{1, 2, \dagger}$, Linjie Yang$^{1}$, Chen Chen$^{2}$, Mubarak Shah$^{2}$, Xiaohui Shen$^{1}$, Heng Wang$^{1}$ \\
$^{1}$ ByteDance $^{2}$Center for Research in Computer Vision, University of Central Florida \\
{\tt\small \{sijiezhu,linjie.yang,heng.wang,shenxiaohui.kevin\}@bytedance.com, \{chen.chen,shah\}@crcv.ucf.edu}}

\maketitle

\let\thefootnote\relax\footnotetext{$\dagger$ This work was done during the first author’s internship at ByteDance}
\begin{abstract}
Visual Place Recognition (VPR) estimates the location of query images by matching them with images in a reference database. Conventional methods generally adopt aggregated CNN features for global retrieval and RANSAC-based geometric verification for reranking. However, RANSAC only employs geometric information but ignores other possible information that could be useful for reranking, \eg local feature correlations, and attention values. In this paper, we propose a unified place recognition framework that handles both retrieval and reranking with a novel transformer model, named $R^{2}$Former. The proposed reranking module takes feature correlation, attention value, and xy coordinates into account, and learns to determine whether the image pair is from the same location. The whole pipeline is end-to-end trainable and the reranking module alone can also be adopted on other CNN or transformer backbones as a generic component. Remarkably, $R^{2}$Former significantly outperforms state-of-the-art methods on major VPR datasets with much less inference time and memory consumption. It also achieves the state-of-the-art on the hold-out MSLS challenge set and could serve as a simple yet strong solution for real-world large-scale applications. Experiments also show vision transformer tokens are comparable and sometimes better than CNN local features on local matching. The code is released at \url{https://github.com/Jeff-Zilence/R2Former}.
\end{abstract}


\section{Introduction}
\label{sec:intro}

\begin{figure}[!htbp]
    \centering
    \includegraphics[width=1.\linewidth]{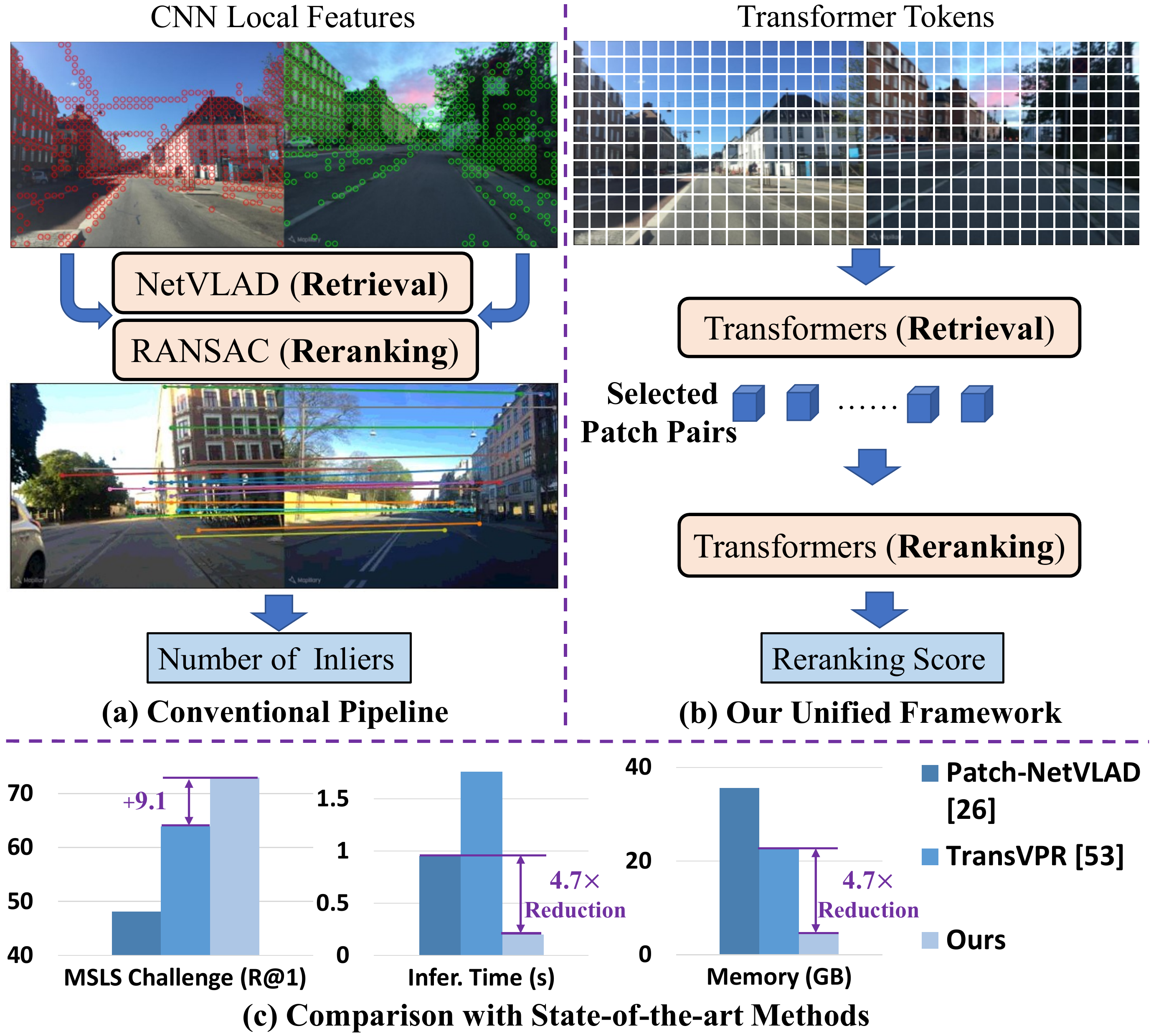}
    \vspace{-0.7cm}
    \caption{An overview of conventional pipeline and our unified framework. Both our global retrieval and reranking modules are end-to-end trainable transformer layers, which achieve state-of-the-art performance with much less computational cost. }
    \label{fig:overview}
    \vspace{-0.3cm}
\end{figure}

Visual Place Recognition (VPR) aims to localize query images from unknown locations by matching them with a set of reference images from known locations. 
It has great potential for robotics \cite{stumm2013probabilistic}, navigation \cite{mirowski2018learning}, autonomous driving \cite{bresson2017simultaneous}, augmented reality (AR) applications. 
Previous works \cite{wang2022transvpr,hausler2021patch} generally formulate VPR as a retrieval problem with two stages, \ie global retrieval, and reranking. The global retrieval stage usually applies aggregation methods (\eg NetVLAD \cite{arandjelovic2016netvlad} and GeM \cite{gem}) on top of CNN (Convolutional Neural Network) 
to retrieve top candidates from a large reference database. While some works \cite{arandjelovic2016netvlad,berton2022deep} only adopt global retrieval, current state-of-the-art methods \cite{wang2022transvpr,hausler2021patch} conduct reranking (\ie geometric verification with RANSAC \cite{fischler1981random}) on the top-k (\eg $k=100$) candidates to further confirm the matches, typically leading to a significant performance boost. \textit{However, geometric information is not the only information that could be useful for reranking, and task-relevant information could be learned with a data-driven module to further boost performance.} Besides, the current reranking process requires a relatively large inference time and memory footprint (typically over 1s and 1MB per image), which cannot scale to real-world applications with large QPS (Queries Per Second) and reference database ($>1$M images). 

Recently vision transformer \cite{vit} has achieved significant performance on a wide range of vision tasks, and has been shown to have a high potential for VPR \cite{wang2022transvpr} (Sec. \ref{sec:related}). However, the predominant local features \cite{hausler2021patch,wang2022transvpr} are still based on CNN backbones, due to the built-in feature locality with limited receptive fields. Although vision transformer \cite{vit} considers each patch as an input token and naturally encodes local information, the local information might be overwritten with global information by the strong global correlation between all tokens in every layer. \textit{Therefore, it is still unclear how vision transformer tokens perform in local matching as compared to CNN local features in this field.} 

In this paper, we integrate the global retrieval and reranking into one unified framework employing only transformers (Fig. \ref{fig:overview}), abbreviated as $R^{2}$Former, which is simple, efficient, and effective. The global retrieval is tackled based on the class token without additional aggregation modules \cite{arandjelovic2016netvlad, gem} and the other image tokens are adopted as local features. Different from geometric verification which focuses on geometric information between local feature pairs, we feed the correlation between the tokens (local features) of a pair of images, xy coordinates of tokens, and their attention information to transformer modules, so that the module can learn task-relevant information that could be useful for reranking. The global retrieval and reranking parts can be either trained in an end-to-end manner or tuned alternatively with a more stable convergence. The proposed reranking module can also be adopted on other CNN or transformer backbones, and \textit{our comparison shows that vision transformer tokens are comparable to CNN local features in terms of reranking performance (Table \ref{tab:ablation-backbone}).} 

Without bells and whistles, the proposed method outperforms both retrieval-only and retrieval+reranking state-of-the-art methods on a wide range of VPR datasets.
The proposed method follows a very efficient design and applies linear layers for dimension reduction: \ie only $256$ and $500\times 131$ for global and local feature dimensions and only $32$ for transformer dimension of reranking module, thus is significantly faster ($>4.7 \times$ QPS) with much less ($<22\%$) memory consumption than previous methods \cite{hausler2021patch,wang2022transvpr}. Both the global and local features are extracted from the same backbone model only once, and the reranking of top-k candidates is finished with only one forward pass by computing the reranking scores of all candidate pairs in parallel within one batch. The reranking speed can be further boosted by parallel computing on multiple GPUs with $>20\times$ speedup over previous methods \cite{hausler2021patch,wang2022transvpr}. We demonstrate that the proposed reranking module also learns to focus on good local matches like RANSAC \cite{fischler1981random}. We summarize our contributions as follows:
\setlist{nolistsep}
\begin{itemize}[noitemsep,leftmargin=*]
    \item A unified retrieval and reranking framework for place recognition employing pure transformers, which demonstrates that vision transformer tokens are comparable and sometimes better than CNN local features in terms of reranking or local matching.
    \item A novel transformer-based reranking module that learns to attend to the correlation of informative local feature pairs. It can be combined with either CNN or transformer backbones with better performance and efficiency than other reranking methods, \eg RANSAC.
    \item Extensive experiments showing state-of-the-art performance on a wide range of place recognition datasets with significantly less ($<22\%$) inference latency and memory consumption than previous reranking-based methods. 
\end{itemize}
\section{Related Work}
\label{sec:related}
\noindent\textbf{Visual Place Recognition.} Visual place recognition (VPR) \cite{angeli2008fast, stumm2013probabilistic , zamir2014image, 7339473, 9336674, ZHANG2021107760, ijcai2021p603} is traditionally addressed with nearest neighbor search on aggregated \cite{jegou2011aggregating} hand-crafted features \cite{csurka2004visual,lowe1999object}. The current predominant methods \cite{berton2022deep,berton2022rethinking,hausler2021patch,wang2022transvpr,berton2021viewpoint,arandjelovic2016netvlad} are based on CNN \cite{vgg,he2016deep} feature extractors with trainable aggregation layer, \eg NetVLAD \cite{arandjelovic2016netvlad}, CRN \cite{jin2017learned}, or light-weighted pooling layer, \eg GeM \cite{gem}, R-MAC \cite{gordo2017end}. Numerous works \cite{berton2021adaptive,gadd2020look,garg2021seqnet,ge2020self,liu2019stochastic,wang2019attention,warburg2020mapillary} follow NetVLAD \cite{arandjelovic2016netvlad} to further improve the global representation for retrieval. Recently, Berton \cite{berton2022deep} \etal introduce a benchmark for VPR and implement a wide range of global-retrieval-based methods in the same framework. CosPlace \cite{berton2022rethinking} \etal propose an extremely large dataset and formulate the problem differently as classification using orientation information.  

In addition to global retrieval, recent state-of-the-art VPR methods \cite{hausler2021patch,wang2022transvpr} apply reranking on the top retrieved candidates. Patch-NetVLAD \cite{hausler2021patch} adopts NetVLAD \cite{arandjelovic2016netvlad} model as their backbone for global retrieval and applies RANSAC \cite{fischler1981random} based geometric verification on multi-scale patch descriptors. TransVPR \cite{wang2022transvpr} adds multiple transformer layers on the top of CNN backbones to extract both global and local features. An additional attention module is then deployed to select important local features as the input of RANSAC \cite{fischler1981random}. Other local matching methods from related tasks, \eg SuperGlue \cite{detone2018superpoint,sarlin2020superglue} and DELG \cite{cao2020unifying}, are also evaluated for VPR. However, they both have a lower accuracy with a much slower inference speed \cite{wang2022transvpr}. Recently, there are several recent learning-based reranking methods from related tasks, \eg landmark matching \cite{rrt,lee2022cvnet} and local matching \cite{jiang2021cotr}. While they either fail to generalize \cite{rrt,lee2022cvnet} on VPR (Table \ref{tab:ablation-rerank}) or have a very different problem setting \cite{jiang2021cotr}, \ie computing point-to-point correspondence between two given images without global retrieval. \textit{To summarize, current state-of-the-art VPR methods rely on RANSAC, and methods from related tasks do not generalize well when directly applied to VPR. Our method takes one step forward to integrate retrieval and reranking into one unified VPR framework with pure transformers.}  \\
\begin{figure*}
    \centering
    \includegraphics[width=0.88\linewidth]{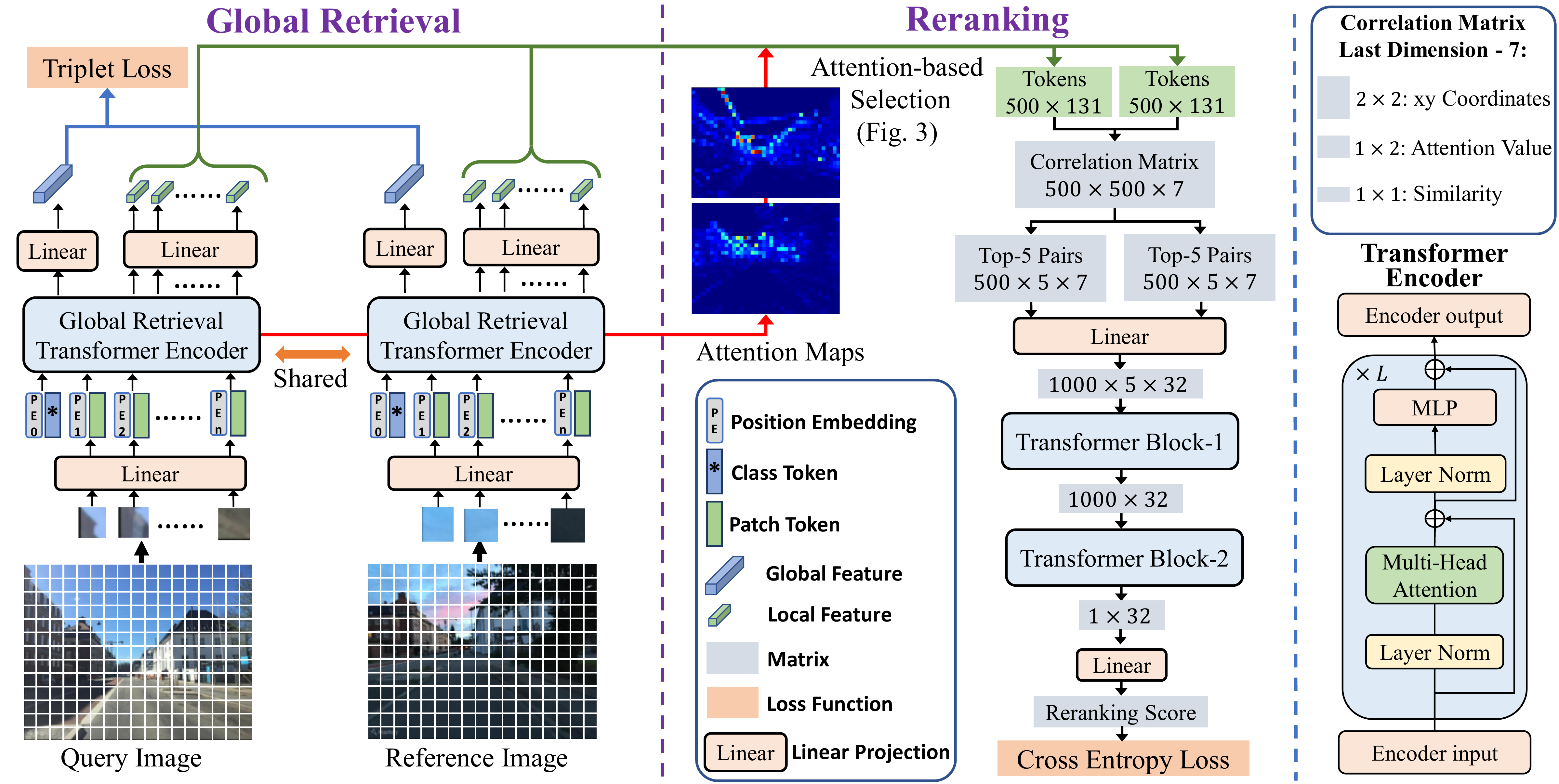}
    \vspace{-0.3cm}
    \caption{An overview of the proposed framework. 
    Both the global and local features are extracted as the class token and patch tokens, followed by dimension reduction using linear layers. The important local features are then selected based on the attention map (Fig. \ref{fig:attention}) and fed into the reranking module. Both the global retrieval and reranking modules consist of only transformer and linear layers.}
    \vspace{-0.3cm}
    \label{fig:architecture}
\end{figure*}
\noindent\textbf{Vision Transformer.} Transformer \cite{vaswani2017attention} is first proposed for NLP (Natural Language Processing) tasks as a generic architecture without inductive bias. It is recently introduced to vision tasks as vision transformer \cite{vit} (ViT) by simply considering each image patch as a token. The vanilla ViT \cite{vit} requires large-scale training datasets (\eg ImageNet-21k \cite{deng2009imagenet}) to achieve comparable results as CNN \cite{he2016deep}, and Deit \cite{deit} proposes a data-efficient training strategy for ViT which outperforms CNN on standard ImageNet-1k. The recent VG benchmark \cite{berton2022deep} adopts vanilla ViT without modifying the input resolution which could be suboptimal, while it already shows competitive performance on global retrieval. TransVPR \cite{wang2022transvpr} adopts multiple transformer layers, while the backbone feature extractor is still based on CNN. \textit{We hypothesize the power of vision transformer is not fully exploited for VPR and it is interesting to know how vision transformer tokens perform on reranking/local matching as compared with the predominant CNN local feature.}

\section{$R^{2}$Former}
We first formulate the problem and training objective of the proposed method in Sec. \ref{sec:formutation}. Then we describe global retrieval and reranking stages in Sec. \ref{sec:retrieval} and Sec. \ref{sec:rerank} respectively. Fig. \ref{fig:architecture} shows an overview of $R^{2}$Former.
\label{sec:method}

\subsection{Problem Formulation and Training Objective}
\label{sec:formutation}
The proposed framework consists of two stages, \ie global retrieval, and reranking. Given a set of query images $\{I_{q}\}$ and reference images $\{I_{r}\}$, the objective of global retrieval is to learn an embedding space in which each query $I_{q}$ is close to its corresponding positive reference image $I_{r}$. During training, reference images from the same location as the query image are defined as positive samples, and the typical threshold is set as $10$ meters. We follow the common practice of previous works \cite{arandjelovic2016netvlad,berton2022deep} to find the nearest reference image for each query in the embedding space as the final positive sample. Other reference images with distances greater than $25$ meters are considered as negative samples. Partial negative mining \cite{berton2022deep} is conducted to select the hardest negative samples from a random subset. We denote the global embedding features of query, positive samples, and negative samples as $E_{q}, E_{p}, E_{n}$ and the global retrieval loss is trained with margin triplet loss:
\begin{equation}
\small
\mathcal{L}_{retrieval} = max(||E_{q}-E_{p}||^{2}-||E_{q}-E_{n}||^{2}+m,0).
    \label{eq:retrieval}
\end{equation}
Here $||.||^{2}$ denotes squared L2 norm and $m$ is the margin. \\
\indent The reranking module takes the local features of two images as input and generates two-logit scores $\mathbb{L}$ as the output of a binary classification, representing the likelihoods for True or False matches. We feed both positive and negative query-reference pairs to the reranking module during training and the cross entropy ($CE$) loss is formulated as:
\begin{equation}
\small
    \mathcal{L}_{reranking} = CE(Softmax(\mathbb{L}_{qr}), \mathbb{I}_{qr}).
    \label{eq:reranking}
\end{equation}
$\mathbb{L}$ and $ \mathbb{I}$ denote the logits scores and ground-truth labels for the query-reference pairs. Although the partial negative mining \cite{berton2022deep} is shown to perform better than full negative mining for global retrieval in \cite{berton2022deep}, the objective of the reranking module is to distinguish top-k retrieved candidates which are harder than partial negative samples \cite{berton2022deep}. To make sure the reranking module sees top-k hardest samples during training, we first freeze the global retrieval module and train the reranking module with randomly selected negative samples from the top-k hardest samples of the full database. The retrieval and reranking modules are then finetuned together with partial negative mining for better performance. Details are provided in Sec. \ref{sec:implement}.

\subsection{Global Retrieval Module}
\label{sec:retrieval}
We describe the components of our vision transformer backbone for VPR and how the final global/local features are generated. As shown in Fig. \ref{fig:architecture}, the query and reference images share the same backbone transformer encoder and there is no additional aggregation \cite{arandjelovic2016netvlad, gem} or key-point \cite{detone2018superpoint} extraction module for global and local feature generation. In other words, all the features and intermediate data are directly generated using only transformers. \\
\textbf{ViT for Place Recognition.} An input image $I \in \mathbb{R}^{h\times w \times c}$ is first divided into small $p\times p$ patches ($p=16$ by default) and converted into a number of tokens $T \in \mathbb{R}^{n\times d}$ by linear projection as the input of transformer encoders. Here $n,d$ denote the number and dimension of tokens, and $h, w, c$ denote the height, width, and number of channels in input images. In addition to the $n$ patch tokens, ViT \cite{vit} adds an additional learnable class token to aggregate classification information from each layer, which serves as a simple alternative for feature aggregation. Then we adopt the learnable position embedding $PE \in \mathbb{R}^{(n+1)\times d}$ from ViT \cite{vit} and add it to each token to provide positional information. \textit{Different from previous work \cite{berton2022deep} using the fixed training resolution of ViT \cite{vit}, we conduct 2D interpolation on the positional embedding so that the input resolution can be arbitrary and we use the most widely used resolution ($640\times 480$) for our method.} \\
\textbf{Global Attention.} On the bottom right of Fig. \ref{fig:architecture}, we provide the detailed inner architecture of the transformer encoder, which has $L$ cascaded basic transformer \cite{vaswani2017attention} layers. The key component is the multi-head attention module, which adopts three linear projections to convert the input token into query, key and value, denoted as $Q, K, V$ with dimension $d$. The basic attention is computed as $softmax(QK^{\mathsf{T}}/d)V$ ($\mathsf{T}$ means transpose), and a multi-head attention module performs this attention procedure in parallel for multiple heads with their final outputs concatenated. \textit{The multi-head attention module can model stronger global correlation than CNN with a limited receptive field, which helps information aggregation for global representation. However, it might reduce the locality of each token which could be harmful to local matching, we demonstrate that transformer tokens perform well as local features compared to CNN as shown in Table \ref{tab:ablation-backbone}.}\\
\textbf{Dimension Reduction.} We feed the output class token from the last transformer layer into a linear layer to generate the global feature with only $256$ dimensions. The local features are generated by applying a linear head on the output patch tokens of the penultimate transformer layer and the dimension is reduced to $128$ to achieve a small memory consumption. Both global and local features are L2 normalized after reduction. \textit{The dimension reduction ensures a much smaller (Table \ref{tab:cost}) total feature dimensions than previous reranking-based methods.} 

\subsection{Reranking Transformer Module}
\label{sec:rerank}
\noindent We describe the attention-based local feature selection and the workflow inside our reranking transformer module.\\
\textbf{Attention-based Selection.} For large-scale scenarios, local features occupy the majority of memory, thus reducing the number of local features is important for real-world deployment. Previous works \cite{arandjelovic2016netvlad,hausler2021patch,wang2022transvpr} usually use $640\times 480$ as image resolution, resulting in 1200 patch tokens. However, only a small portion of them are informative to determine the location of the image, \eg buildings, trees, roads, \etc. Previous works either leverage all the local features from multiple scales \cite{hausler2021patch} or adopt additional modules \cite{wang2022transvpr} to generate attention maps, which filter local features with a certain threshold during local matching. \textit{On the contrary, we leverage the natural attention map within our transformer module and only save a fixed number of local features (\eg $500$) with top attention values, resulting in a much lower memory cost and simpler extraction pipeline.} \\
\begin{figure}[tbp]
    \centering
    \includegraphics[width=0.9\linewidth]{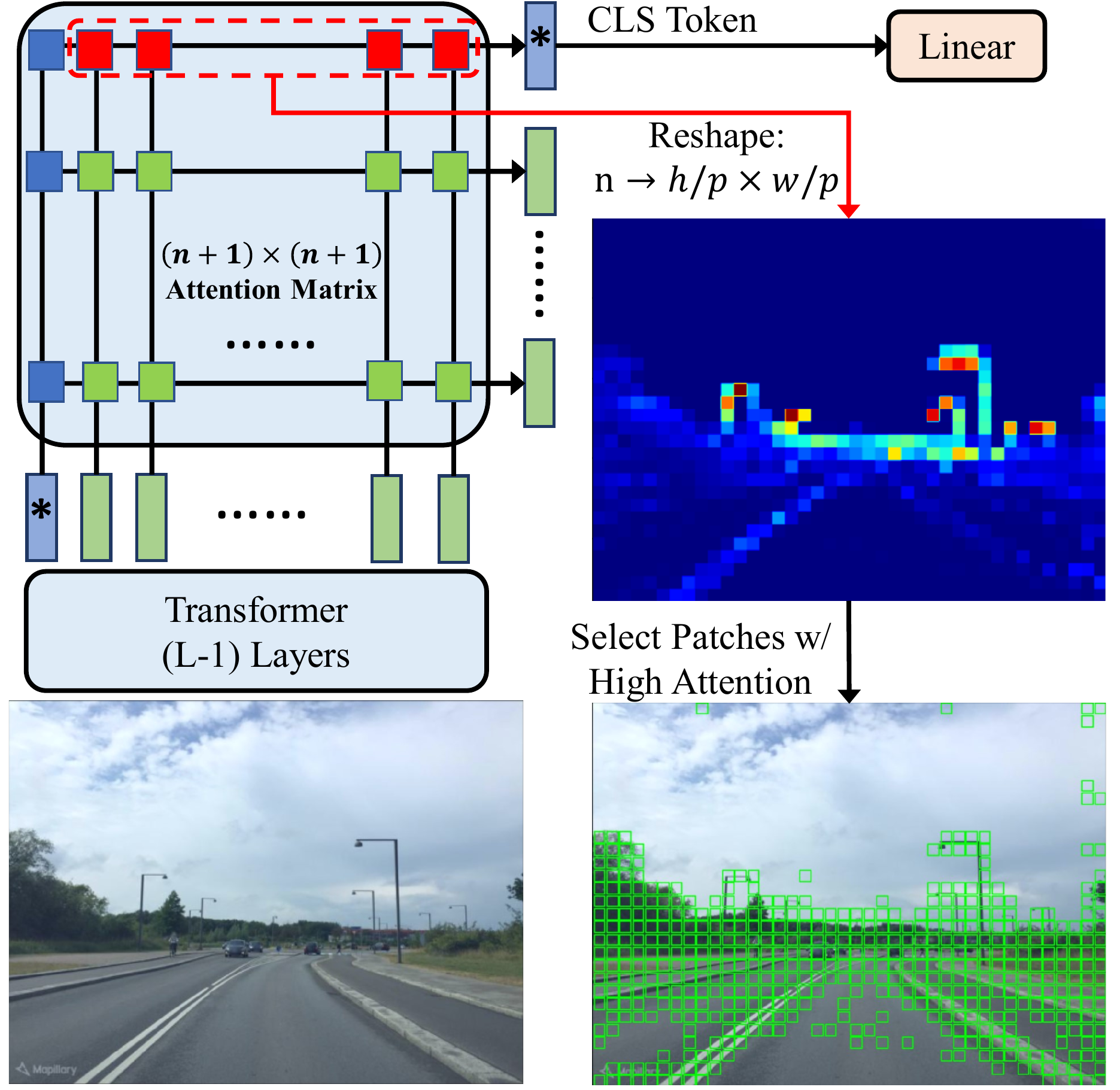}
    \vspace{-0.1cm}
    \caption{Illustration of attention generation and token selection.}
    \label{fig:attention}
    \vspace{-0.2cm}
\end{figure}
The $(n+1)\times (n+1)$ attention matrix of the last transformer layer in Fig. \ref{fig:attention} is formulated as $softmax(QK^{\mathsf{T}}/d)$ in Sec. \ref{sec:retrieval}. It represents the contribution from each input token to each output token. Since the output class (CLS) token is the only one that connects to the global embedding feature, the CLS token output channel of the matrix ($(n+1)$-dimensional vector for $n$ patches and CLS token) represents the contribution from input tokens to the global feature, which corresponds to the importance of each patch. We reshape the $n$-dimensional vector corresponding to $n$ patches as $h/p\times w/p$ attention map to sort all the tokens and we select the top-500 tokens that are likely to cover the informative regions (bottom right of Fig. \ref{fig:attention}). The attention value and x,y coordinates are saved along with each local feature, resulting in $128+3=131$ dimensions. \\
\noindent\textbf{Correlation-based Reranking.} RANSAC-based \cite{fischler1981random} geometric verification only leverages the top-1 matched pairs for all local features, but ignores other important information that could be useful for reranking, \ie the correlation, and attention value. Local feature pairs with higher correlation/similarity have higher probabilities to be correct local matches, and an image pair with more correct local matches is more likely to represent the same place. Also, high attention values of the local patches could indicate the importance of the local feature pairs. Therefore, our reranking module is designed to maintain the correlation, attention, and positional information for all the local feature pairs in a correlation matrix, resulting in 7 dimensions (Fig. \ref{fig:architecture}) denoted as $(x, y, A, x', y', A', S)$. $x, y, x', y'$ denote the coordinates of the two patches in the query and reference images. $A, A'$ denotes the attention values of the two patches. $S$ is the cosine similarity between the $128$-dimensional local features. Since each image has 500 selected features/tokens, there are $500 \times 500$ pairs, resulting in a $500\times 500 \times 7$ correlation matrix. \textit{The correlation matrix contains the major information for all the feature pairs and allows the model to learn whatever is useful to determine whether the two images are from the same location.}

To reduce the computation, we select the 5 nearest neighbors of each token in the feature space to produce two $500\times 5 \times 7$ matrices, as the other feature pairs with large distances are likely to be wrong matches. They are concatenated together and fed to a linear layer, resulting in a $1000 \times 5\times 32$ matrix. \textit{We then leverage the strong global correlation modeling of transformer \cite{vaswani2017attention} to aggregate the large matrix as a reranking score to determine whether the input pair is a correct match.} First, we adopt ``Transformer Block-1" to extract important information from the top-5 pairs as one output class token. Then ``Transformer Block-2" extracts aggregates the information from the 1000 tokens as a single $32$-dimension vector (the class token). The two transformer blocks are multiple transformer layers with linear projection and standard Sinusoidal positional embedding \cite{vaswani2017attention}. Finally, the vector is converted into $2$ channels by a linear head as a binary classification (\ie True vs False). 


\section{Experiment}
\label{sec:experiment}

\subsection{Datasets and Evaluation Metrics}
We train our model on MSLS (Mapillary Street-Level Sequences) \cite{warburg2020mapillary} dataset, which covers a wide range of real-world scenarios for VPR, \eg different cities, viewpoint variation, day/night, and season changes. The performance in urban scenarios (\eg Pitts30k \cite{pitts30k}, Tokyo24/7 \cite{tokyo247}) can be further improved by finetuning on Pitts30k. For evaluation, we use standard train/val/test split \cite{warburg2020mapillary,hausler2021patch,wang2022transvpr,berton2022deep} on major datasets, including MSLS Val \cite{warburg2020mapillary}, MSLS Challenge \cite{warburg2020mapillary}, Pitts30k \cite{pitts30k}, Tokyo 24/7 \cite{tokyo247}, R-SF (Revisited San Francisco) \cite{chen2011city, li2012worldwide, sattler2017large, berton2022deep}, St Lucia \cite{milford2008mapping}. The MSLS Challenge \cite{warburg2020mapillary} is a hold-out set whose labels are not released, but researchers can submit the predictions on their challenge server to get the performance. For the other datasets with location labels, we follow previous works \cite{hausler2021patch,wang2022transvpr,berton2022deep} to use 25 meters as the threshold for correct localization and report recall@k (k=1,5,10) as evaluation metrics. We also report detailed computational costs, including inference time for extraction/retrieval/matching, feature dimensions, GFLOPs, and memory footprint (Table \ref{tab:cost}).  

\subsection{Implementation Details}
\label{sec:implement}
The proposed method is implemented using PyTorch \cite{paszke2019pytorch}. The models are trained on 8 Tesla-V100 GPUs. All images are resized to $640\times 480$ for training and evaluation. The reranking is conducted on top-100 candidates and we set margin $m=0.1$. We use ViT-S \cite{vit} (12 layers with 384 dimensions) architecture as our default backbone and the module is initialized with off-the-shelf pre-trained weights \cite{deit} on ImageNet-1k \cite{deng2009imagenet}. The transformers in the reranking module use a small dimension of $32$ with only $2$ and $6$ layers for Transformer Block-1 and Block-2 respectively, thus resulting in a very small computational cost.  

The global retrieval and the reranking module can be trained jointly in an end-to-end manner, but we train them separately (Sec. \ref{sec:formutation}) by default to achieve a stable convergence and better accuracy. The global retrieval module is trained following common practice \cite{berton2022deep}. 
The reranking module is trained from scratch following the standard pipeline of training transformers, \ie AdamW \cite{loshchilov2017decoupled} optimizer with cosine learning rate schedule \cite{loshchilov2016sgdr}. The initial learning rate is 0.0005 with 64 triplets per batch. Each triplet only samples one hard negative reference image which is randomly selected from the top-100 hardest samples from the MSLS training set of the corresponding query. Given that the global embedding features do not change during this stage, the global hard negative mining only needs to be conducted once and can be efficiently computed on GPU in 3 hours. We precompute the hard negative list of all queries for the training of the reranking module, which is more efficient than the full mining implementation of the VG benchmark \cite{berton2022deep}. The reranking module is trained for 50 epochs and we select the model with the highest recall@5 on the validation set. The global retrieval and reranking module are then finetuned together with partial negative mining \cite{berton2022deep}.
More details are included in \textbf{supplementary material}.


\begin{table*}[!htbp]
    \centering
    \resizebox{\linewidth}{!}{
    \begin{tabular}{l c c c c c c c c c c c c c c c }
    \hline
    
    \hline
     \multirow{2}{*}{}  & \multicolumn{3}{c}{MSLS Val  \cite{warburg2020mapillary}} & & \multicolumn{3}{c}{MSLS Challenge  \cite{warburg2020mapillary}} & & \multicolumn{3}{c}{Pitts30k \cite{pitts30k}} &  & \multicolumn{3}{c}{Tokyo 24/7 \cite{tokyo247}} \\
     \cline{2-4}  \cline{6-8} \cline{10-12} \cline{14-16}
         & R@1 & R@5 & R@10 & & R@1 & R@5 & R@10 & & R@1 & R@5 & R@10 & & R@1 & R@5 & R@10  \\
    \hline
    NetVLAD \cite{arandjelovic2016netvlad} & 60.8 & 74.3 & 79.5 & & 35.1 & 47.4 & 51.7 & & 81.9 & 91.2 & 93.7 & & 64.8 & 78.4 & 81.6 \\
    SFRS \cite{ge2020self}  & 69.2 & 80.3 & 83.1 & & 41.5 & 52.0 & 56.3 & & 89.4 & 94.7 & 95.9 & & 85.4 & 91.1 & \textbf{93.3} \\
    SP-SuperGlue \cite{detone2018superpoint,sarlin2020superglue} & 78.1 & 81.9 & 84.3 & & 50.6 & 56.9 & 58.3 & & 87.2 & 94.8 & \textbf{96.4} & & 88.2 & 90.2 & 90.2  \\
    Patch-NetVLAD \cite{hausler2021patch} & 79.5 & 86.2 & 87.7 & & 48.1 & 57.6 & 60.5 & & 88.7 & 94.5 & 95.9 & & 86.0 & 88.6 & 90.5 \\
    TransVPR \cite{wang2022transvpr} & 86.8  & 91.2  & 92.4 & & 63.9 & 74.0 & 77.5 & & 89.0 & 94.9 & 96.2 & & 79.0 & 82.2 & 85.1 \\
    \hline
    Ours & \textbf{89.7} & \textbf{95.0} & \textbf{96.2} & & \textbf{73.0} & \textbf{85.9} & \textbf{88.8} & &  \textbf{91.1} & \textbf{95.2} & 96.3 &  &  \textbf{88.6} & \textbf{91.4} & 91.7 \\
    \hline
    
    \hline
    \end{tabular}}
    \vspace{-0.2cm}
    \caption{Comparison of our method with previous state-of-the-art results on major VPR datasets. Our model is trained on MSLS and tested on MSLS Val and Challenge set. Our model is further finetuned on Pitts30k for urban scenarios, \ie Pitts30k, Tokyo 24/7.} 
    \label{tab:main}
    \vspace{-0.2cm}
\end{table*}

\begin{table*}[!htbp]
\small
    \centering
    \begin{tabular}{l c c c c c c c c}
    \hline
    
    \hline
    & \multirow{2}{*}{Features Dim $\downarrow$} & \multirow{2}{*}{GFLOPs $\downarrow$} & \multicolumn{6}{c}{Trained on MSLS - R@1}\\ 
    \cline{4-9}
        &  &  & \footnotesize{Pitts30k} & \footnotesize{MSLS Val} & \footnotesize{MSLS Chall.} & \footnotesize{Tok. 24/7} & \footnotesize{R-SF}  & \footnotesize{St Lucia}  \\ 
    \hline 
    ResNet101 + GeM \cite{gem, berton2022deep}  & 1024 & 86.29 & 77.2 & $77.0^{*}$ & 55.5 & 51.0 & 46.9  & 91.6\\ 
    ResNet101 + NetVLAD \cite{arandjelovic2016netvlad, berton2022deep} & 65536 & 86.06 & 80.8 & $81.1^{*}$ & 61.5 & 59.0 & 56.1  & 95.1 \\ 
    CCT384 + NetVLAD \cite{hassani2021escaping, berton2022deep} & 24576 & \textbf{18.53} & 85.1 & $83.8^{*}$ & 61.4 & 70.3 & 65.9  & 98.4 \\ 
    \hline
    Ours w/o Reranking & $ \textbf{256}$ & 25.90  & 76.3 & 79.3 & 56.2 & 45.7 & 47.5  & 94.3\\ 
    Ours & $65500^{\dagger}$ & $48.50^{\dagger}$ & \textbf{88.4} & \textbf{89.7} & \textbf{73.0} & \textbf{72.7} & \textbf{72.1} & \textbf{99.7} \\
    \hline
    
    \hline
    \end{tabular}
    \vspace{-0.2cm}
    \caption{Comparison following the protocol of the recent VG benchmark \cite{berton2022deep} on major datasets in terms of R@1. The models are trained on MSLS and directly tested on all datasets w/o finetuning.  $\dagger$ The dimensions and GFLOPs are computed by adding the numbers of global retrieval and reranking together, \ie $Dim = 256 + 500 \times (128+3)$ and $GFLOPs = 25.90 + 100 \times 0.226$.} 
    \vspace{-0.2cm}
    \label{tab:benchmark}
\end{table*}

\begin{table*}[!htbp]
\small
    \centering
    \begin{tabular}{l c c c c c c c c c}
    \hline
    
    \hline
         & \multicolumn{2}{c}{Feature Dim $\downarrow$} & & \multicolumn{3}{c}{Latency per Query (ms) $\downarrow$} & & \multicolumn{2}{c}{Memory Footprint (GB) $\downarrow$}  \\
         \cline{2-3} \cline{5-7} \cline{9-10}
         & Global & Local & & Extraction & Retrieval & Reranking & & MSLS Val & 1M Images \\
    \hline
    ResNet101 + NetVLAD \cite{arandjelovic2016netvlad, berton2022deep} & 65536 & N/A & & 9.60 & 2.33 & N/A &  & 4.79 & 244.14 \\ 
    Patch-NetVLAD-s \cite{hausler2021patch} & 512 &  $936 \times 512$ & & 9.29 & 0.08 & 952.85 & & 37.60 & 1917.29 \\
    Patch-NetVLAD-p \cite{hausler2021patch} & 4096 & $2826 \times 4096$ & & 9.36 & 0.19 & 8377.17 &  & 908.30 & 46315.85 \\ 
    TransVPR \cite{wang2022transvpr} & $\textbf{256}$ & $1200 \times 256$ & & \textbf{6.20} & \textbf{0.07} & 1757.70 &  & 22.72 & 1158.53 \\
    \hline
     Ours & $\textbf{256}$ & $\textbf{500}\times \textbf{(128+3)}$ & & 8.81 & \textbf{0.07} & \textbf{202.37} & & \textbf{4.79} & \textbf{244.01}\\ 
    \hline
    
    \hline
    \end{tabular}
    \vspace{-0.2cm}
    \caption{Comparison of computational cost in terms of feature dimension, latency, and memory footprint. All the methods are measured on MSLS Val ($18,871$ database images) using the same CPU and GPU (RTX A5000). Reranking is conducted on top-100 candidates. ``Patch-NetVLAD-p" and ``Patch-NetVLAD-s" denote the performance and speed-oriented versions.}
    \label{tab:cost}
    \vspace{-0.2cm}
\end{table*}

\subsection{Comparison with State-of-the-art}

In this section, we compare the proposed method with previous state-of-the-art methods, including NetVLAD \cite{arandjelovic2016netvlad}, SFRS \cite{ge2020self}, SP-SuperGlue \cite{detone2018superpoint,sarlin2020superglue}, Patch-NetVLAD \cite{hausler2021patch}, and TransVPR \cite{wang2022transvpr}. We further compare our method with the best configurations in the recent VG benchmark \cite{berton2022deep}. We also provide a detailed comparison with state-of-the-art methods \cite{berton2022deep, hausler2021patch, wang2022transvpr} on computational cost. Other works \cite{doersch2020crosstransformers, fan2022svt, xu2023transvlad, zhang2023etr, ali2023mixvpr, zhu2022transgeo} on related tasks with different settings are not included. \\
\indent As shown in Table \ref{tab:main}, the proposed method significantly outperforms state-of-the-art methods on  MSLS Val and Challenge set with absolute R@1 \textbf{improvement of $2.9\%$ and $9.1\%$ respectively}. Benefiting from large-scale training, our method also achieves state-of-the-art R@1 on Pitts30k and Tokyo24/7. Small-scale dataset \cite{pitts30k} with weak supervision is not enough to train a data-driven reranking module, and training on one large-scale dataset with good generalization on all datasets is more desirable for real-world deployment. (Details in supp. material.)\\
\indent In Table \ref{tab:benchmark}, we select the best-performing models in the VG benchmark \cite{berton2022deep} and compare them with our method on major datasets. Eynsham \cite{Cummins2009HighlySA} dataset is not included because it has a different camera type (gray-scale) and resolution. 
$*$ denotes reproduced results using the official MSLS \cite{warburg2020mapillary} validation code and pre-trained models \cite{berton2022deep}. Table \ref{tab:benchmark} shows that our model trained on MSLS generalizes well on all the other datasets and significantly outperforms the best models in the VG benchmark \cite{berton2022deep} with comparable computational cost.\\
\begin{table*}[!htbp]
\begin{minipage}[t]{.44\linewidth}
    \centering
    \begin{tabular}{l c c c c }
    \hline
    
    \hline
         & R@1 & R@5 & R@10  \\
    \hline
     No Reranking & 79.3 & 90.8 & 92.6 \\
     RANSAC \cite{fischler1981random} & 84.9 & 93.0 &  94.5 \\ 
     RRT \cite{rrt} & 81.2 & 91.9 & 93.1 \\
     CVNet \cite{lee2022correlation} & 73.4 &  86.8 & 91.4 \\
     \hline
     Ours &  \textbf{89.7} & \textbf{95.0} & \textbf{96.2} \\
    \hline
    
    \hline
    \end{tabular}
    \vspace{-0.2cm}
    \captionof{table}{Comparison with different reranking methods (top-100 candidates reranked) using our backbone on MSLS Val.}
    \label{tab:ablation-rerank}
\end{minipage}\hspace{0.4cm}%
\begin{minipage}[t]{.53\linewidth}
    \centering
    \begin{tabular}{l c c c c }
    \hline
    
    \hline
         & R@1 & R@5 & R@10  \\
    \hline
    No Reranking & 79.3 & 90.8 & 92.6 \\
    Reranking & 86.6 & 94.1 & 95.0 \\
    Reranking + Mining & 88.4 & 93.4 & 94.9 \\
    Reranking + Mining + Finetune & \textbf{89.7} & \textbf{95.0} & \textbf{96.2}  \\
    End-to-end Training & 87.3 & 93.5  & 95.4 & \\
    \hline
    
    \hline
    \end{tabular}
    \vspace{-0.2cm}
    \captionof{table}{Ablation study on training strategy. We freeze the global retrieval module by default when training the reranking module.} 
    \label{tab:ablation-training}
\end{minipage}
\vspace{-0.3cm}
\end{table*}
In Table \ref{tab:cost}, we compare the computational cost of our method with previous global retrieval methods and retrieval+reranking methods. Apparently, the inference latency of extraction and retrieval is negligible as compared to the reranking latency of previous works. Our method is significantly faster than all the other reranking-based methods because our reranking process is only one network forward pass on GPU and can be easily accelerated by parallel computing, \eg reranking on 4 GPUs takes only $47.17$ ms per query ($20.2 \times$ speedup over previous reranking-based methods \cite{hausler2021patch,wang2022transvpr}). 
However, it is not straightforward to accelerate RANSAC-based geometric verification with multi-processing. Previous works \cite{hausler2021patch, wang2022transvpr} generally follow the standard implementation of OpenCV \cite{opencv_library}. In addition, the local feature dimension of our method is the smallest among all the reranking-based methods, specifically, we only save $500$ important local features with very small dimensions ($128+3$) for each image. In total, it only needs $4.79$ GB of memory for the MSLS validation set which is much smaller ($189\times$ reduction) than $908.3$ GB of Patch-NetVLAD-p. Even for datasets with over 1M reference images (\eg R-SF dataset), the memory footprint is only 244.01 GB which can fit in a typical server or high-end desktop. It can be further reduced to $122$ GB by using float16 instead of float32, with which we observe no performance drop on R-SF. However, other reranking-based methods typically require much larger ($> 4\times$) memory, which cannot scale to large-scale real-world scenarios with 1M reference images. Given the unified design, outstanding performance, strong efficiency, and scalability of our method, it could serve as a well-balanced solution for real-world large-scale applications.

\begin{table}[tbp]
\small
    \centering
    \begin{tabular}{l l c c c }
    \hline
    
    \hline
         & Architecture & R@1 & R@5 & R@10 \\
    \hline
    \multirow{3}{1.5cm}{Ours w/o\\ Reranking}  & ViT-Small  & 79.3 & 90.8 & 92.6\\ 
                                            & ResNet50 + GeM  & 79.6 & 90.9 &  92.6 \\
                                            & ViT-Base & 84.9 & 92.7 & 94.5 \\
    \hline
    \multirow{3}{1.5cm}{Ours w/\\ RANSAC} &
    ViT-Small     & 84.9 & 93.0 &  94.5 \\ 
      & ResNet50 + GeM & 84.3 &  91.4 & 93.0 \\
      & ViT-Base & 87.0 & 93.0 & 94.6 \\
    \hline
    \multirow{3}{1.5cm}{Ours} 
      & ViT-Small    & 89.7 &  95.0 & 96.2 \\ 
      & ResNet50 + GeM & 88.4 & 93.6 & 95.3 \\
      & ViT-Base & \textbf{90.0} & \textbf{95.1} & \textbf{96.9} \\
    \hline
    
    \hline
    \end{tabular}
    \vspace{-0.2cm}
    \caption{Ablation study on MSLS Val with different backbone architectures, including ResNet50+GeM \cite{he2016deep,gem}, ViT Small\cite{vit}, ViT Base\cite{vit}. Transformer tokens perform on par with CNN local features in terms of reranking/local matching.} 
    \vspace{-0.2cm}
    \label{tab:ablation-backbone}
\end{table}

\subsection{Ablation Study}
\noindent\textbf{Reranking Methods.} In Table \ref{tab:ablation-rerank}, we adopt the same global and local features from our default backbone and compare different reranking methods. ``No Reranking" denotes our model with only global retrieval. ``RANSAC" \cite{fischler1981random} follows the pipeline of state-of-the-art methods \cite{hausler2021patch, wang2022transvpr} using $1.5 \times$ of the patch size. ``RRT" \cite{rrt} is not proposed for VPR, nevertheless, we train it on the top of our backbone for comparison. ``CVNet" \cite{lee2022cvnet} does not provide the training code, we thus adopt their pre-trained model trained on Google Landmark \cite{weyand2020google}. Remarkably, our reranking module significantly outperforms RANSAC \cite{fischler1981random} based geometric verification using much less ($<22\%$) inference time (Table \ref{tab:cost}). It also outperforms two recent state-of-the-art reranking methods on landmark retrieval, indicating the superiority of our reranking module. \\
\begin{figure*}[!htbp]
    \centering
    \includegraphics[width=1.\linewidth]{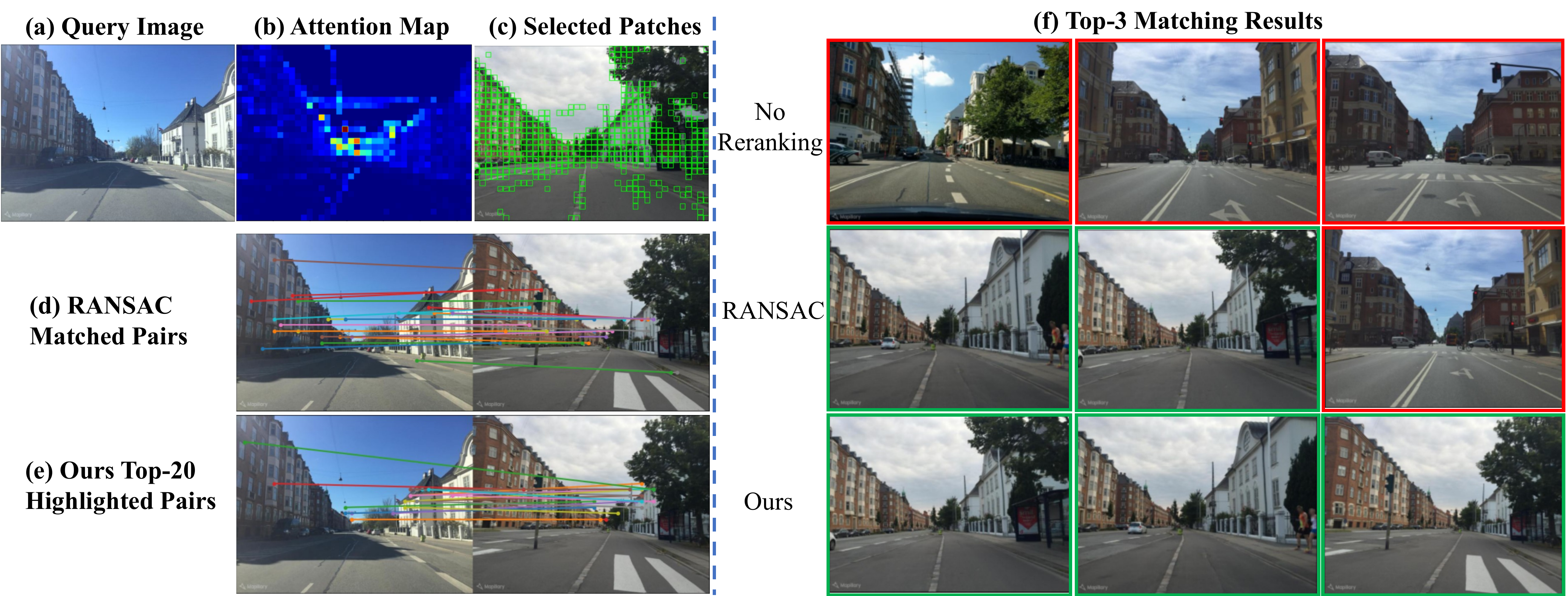}
    \vspace{-0.5cm}
    \caption{Visualization of attention map and matched pairs. Green and red boxes in (f) indicate correct and wrong predictions. }
    \vspace{-0.2cm}
    \label{fig:case}
\end{figure*}
\noindent\textbf{Training Strategies.} In Table \ref{tab:ablation-training}, we compare our method based on different training strategies. ``No Reranking" uses only global retrieval, and ``Reranking" denotes training reranking module with partial negative mining on a fixed global retrieval module. ``Reranking+Mining" means training the reranking module with our global negative mining by randomly selecting one negative sample from the 100 global hardest samples. ``Reranking+Mining+Finetune" further finetunes retrieval and reranking module together. ``End-to-end Training" directly trains global retrieval and reranking module from the beginning in an end-to-end manner with partial negative mining. All the strategies with reranking outperform previous state-of-the-art methods on R@5, and training the global retrieval and reranking module separately with a carefully designed strategy achieves the best performance. The results indicate that global hard negative mining is helpful for training the reranking module. End-to-end training is also a good trade-off between simplicity and performance. \\
\begin{table}[tbp]
\small
    \centering
    \begin{tabular}{l c c c c }
    \hline
    
    \hline
         &  R@1 & R@5 & R@10  \\
    \hline
     No Reranking & 79.3 & 90.8 & 92.6 \\ 
     Reranking & \textbf{86.6} & \textbf{94.1} & \textbf{95.0} \\
     \hline
     \multicolumn{4}{c}{Reranking Module}\\
     \hline
     Remove Positional Embedding & 86.4 & 93.2 & 94.7 \\ 
     Remove Transformer Block-1 &  85.5 & 93.8 & 95.0 \\
     Remove Transformer Block-2 & 81.9 & 92.8 & 94.6\\ 
     Replace Top-5 w/ Top-1 & 86.6 & 93.8 & 94.7 \\
     \hline
     \multicolumn{4}{c}{Reranking Input}\\
     \hline
     Remove Attention Selection & 83.6 & 92.8 & 94.1 \\
     Remove xy Coordinates Value & 85.0 & 93.5 & 94.7 \\
     Remove Attention Value & 86.1 & 93.5 & 94.3 \\
     Remove Correlation Value & 31.8 & 59.2 & 72.8 \\
    \hline
    
    \hline
    \end{tabular}
    \vspace{-0.2cm}
    \caption{Ablation study on model components. We freeze the global retrieval module so that the global and local features are the same for different ablations. All ablations are trained with partial negative mining \cite{berton2022deep}. The components are defined in Sec. \ref{sec:method}.}
    \vspace{-0.3cm}
    \label{tab:ablation-component}
\end{table}
\noindent\textbf{Different Backbones.} In Table \ref{tab:ablation-backbone}, we combine our reranking module with different backbone retrieval modules, including ViT-Small (default), ResNet50+GeM \cite{he2016deep,gem} and ViT-Base. We follow \cite{berton2022deep} to add L2 normalization before the pooling layer but do not freeze or remove any convolutional layer. The attention map of ResNet50+GeM is generated following CAM \cite{zhou2016cvpr}. The ResNet50+GeM model achieves competitive results for both global retrieval and reranking. It also significantly outperforms previous works \cite{wang2022transvpr,hausler2021patch}, indicating that our reranking module is a generic component for both CNN and transformer backbone architectures. With similar global retrieval performance and the same reranking method (RANSAC or Ours), the ViT-Small model achieves slightly better performance than ResNet50+GeM, which means the transformer token is a good alternative for local matching as compared with CNN local feature. Furthermore, we adopt a larger backbone ViT-Base, which achieves significant performance improvement on global retrieval and RANSAC-based reranking over ViT-Small. However, the overall performance of our ViT-Base model has very small performance improvement (mainly on R@10), which indicates a possible bottleneck of the proposed framework.\\
\noindent\textbf{Components Ablation.} In Table \ref{tab:ablation-component}, we use the ``Reranking" configuration in Table \ref{tab:ablation-training} as a baseline and ablate different components to check the importance of each component of our method. For simplicity, we freeze the global retrieval module and all the models are trained with partial negative mining. We have the following observations. (1) The most important components are the Transformer Block-2 and Correlation Value, and removing them causes a significant performance drop. This indicates that the feature correlation contributes most to the final decision. (2) When we replace the attention-based feature selection with random selection, the model becomes unstable and the performance drops moderately. (3) Removing other components, \ie Transformer Block-1, xy Coordinates Value, Attention Value (defined in Sec. \ref{sec:retrieval}), causes a slight performance drop, indicating the effectiveness of these blocks for performance purposes. (4) We observe negligible performance drop when removing positional embedding or replacing top-5 pairs selection with top-1, which means the Sinusoidal positional embedding does not help much and the top-1 pair contains most of the information that is needed for reranking.  \\

\subsection{Visualization}

In Fig. \ref{fig:case}, we show a detailed case where the global retrieval fails on the top-3 results. Fig. \ref{fig:case} (b) and (c) show the attention map and the selected patches, which cover most of the informative regions for reranking. Both RANSAC and our method work well on reranking the top-100 candidates and our method finds more correct reference images in the top-3 results (Fig. \ref{fig:case} (f)). Although our reranking module is not as explainable as RANSAC, the transformer still focuses on specific patch pairs and we plot the top-20 pairs with the highest attention in the reranking module. As shown in Fig. \ref{fig:case} (d) and (e), our reranking also focuses on meaningful pairs and most of them are correct matches. As compared with RANSAC, our matched pairs may not have tight geometric consistency, but they may provide different information for the reranking as adjusted according to the task-relevant training data. 




\section{Conclusion}
\label{sec:conclusion}
We propose a unified retrieval and reranking framework for place recognition with only transformers. We for the first time show that vision transformer tokens are comparable to or even better than CNN local feature in terms of reranking. Our reranking module is generic and can be adopted on other CNN or transformer backbones. Remarkably, it significantly outperforms previous methods on major datasets with much less inference time and memory consumption. Our method can also provide possible matched local pairs like RANSAC and could be improved with geometric modeling in the future. We discuss the limitations and societal impact in the supplementary material.

{\small
\bibliographystyle{ieee_fullname}
\bibliography{egbib}
}
\end{document}